\documentclass{article}
\usepackage{spconf,amsmath,epsfig}
\usepackage{graphicx}
\usepackage{amsfonts,amssymb}
\usepackage{bbm}
\usepackage{textcomp}
\usepackage[colorlinks=true, citecolor={blue}, bookmarks=false, breaklinks]{hyperref}
\usepackage{url}

\usepackage{breakurl}
\usepackage{multirow}
\usepackage[normalem]{ulem}
\useunder{\uline}{\ul}{}
\usepackage{adjustbox}
\usepackage[numbers]{natbib}

\begin{document}\sloppy

\def\x{{\mathbf x}}
\def\L{{\cal L}}

\title{Detecting and Correcting Adversarial Images Using Image Processing Operations}


\name{Huy H. Nguyen$^{\star}$, Minoru Kuribayashi$^{\S}$, Junichi Yamagishi$^{\star\dagger\#}$, and Isao Echizen$^{\star\dagger\ddagger}$}
\address{$^{\star}$The Graduate University for Advanced Studies, SOKENDAI, Kanagawa, Japan\\
	$^{\S}$Graduate School of Natural Science and Technology, Okayama University, Okayama, Japan\\
	$^{\dagger}$National Institute of Informatics, Tokyo, Japan; $^{\ddagger}$The University of Tokyo, Tokyo, Japan\\
	$^{\#}$The University of Edinburgh, Edinburgh, UK\\
	\small{Email: \{nhhuy, jyamagis, iechizen\}@nii.ac.jp; kminoru@okayama-u.ac.jp}}

\maketitle

\begin{abstract}
Deep neural networks (DNNs) have achieved excellent performance on several tasks and have been widely applied in both academia and industry. However, DNNs are vulnerable to adversarial machine learning attacks, in which noise is added to the input to change the network output. We have devised an image-processing-based method to detect adversarial images based on our observation that adversarial noise is reduced after applying these operations while the normal images almost remain unaffected. In addition to detection, this method can be used to restore the adversarial images' original labels, which is crucial to restoring the normal functionalities of DNN-based systems. Testing using an adversarial machine learning database we created for generating several types of attack using images from the ImageNet Large Scale Visual Recognition Challenge database demonstrated the efficiency of our proposed method for both detection and correction.
\end{abstract}


\section{Introduction}
\label{sec:intro}
Although adversarial machine learning is not a new issue in the machine learning community, as it was first discussed in 2004~\cite{dalvi2004adversarial}, it has recently become a major concern due to the advances made in deep learning that have made deep neural networks (DNNs) vulnerable to adversarial attacks~\cite{szegedy2013intriguing}. Besides traditional logical attacks, in which adversarial noise is added to image or audio files, attackers can now create physical adversarial examples~\cite{kurakin2016adversarial, eykholt2018physical, boloor2019simple, schoenherr2019}. When autonomous systems have become mainstream in both the research community and industry, physical adversarial attacks may threaten their safety and reliability. Besides white-box attacks, in which attackers have full knowledge of the inner configuration of the target models, attackers will also be able to perform black-box attacks, which are more likely since attackers need only acquire only the models' outputs~\cite{xu2019adversarial}.

Several countermeasures have been proposed for detecting adversarial examples such as statistical testing~\cite{grosse2017statistical}, directly detecting pixels in input images~\cite{gong2017adversarial}, detecting using features from intermediate layers of the targeted DNN~\cite{metzen2017detecting}, applying statistical analysis to the outputs of the intermediate layers~\cite{li2017adversarial}, and using adaptive noise reduction~\cite{liang2018detecting}. Xu et al. used feature squeezing to reduce the search space available to an adversary which was useful for detection~\cite{xu2017feature}. Besides detection, there have been several attempts to harness the DNN models in training by using distillation~\cite{papernot2016distillation}, obfuscating gradients~\cite{athalye2018obfuscated}, or applying the reverse cross-entropy loss function~\cite{pang2018towards}.
Input preprocessing is another direction to defense against adversarial example attacks~\cite{guo2017countering, prakash2018protecting}, however, their performances are still limited. Several adversarial databases have been independently created for use in evaluating the proposed methods, but detailed guidelines for creating them have not been reported.

In this paper, \textbf{first}, we present the procedure we used for creating an adversarial database using images from the ImageNet Large Scale Visual Recognition Challenge (ILSVRC) 2012 validation set~\cite{ILSVRC15}. We used the FoolBox library~\cite{rauber2017foolbox} to create adversarial images for two types of attack: targeted attacks, which force the outputs to become target labels, and non-targeted attacks, which force the top-5 outputs to not include certain labels~\cite{xu2019adversarial} (discussed in section~\ref{sec:create}). \textbf{Second}, we discuss the measured effects of some image processing operations on both normal and adversarial images given the hypothesis that adversarial noise is reduced, resulting in the classification labels of the adversarial images being changed while the normal images almost remain unaffected (section~\ref{subsec:effects}). This measurement was motivated by the work of Xu et al.~\cite{xu2017feature}, Guo et al.~\cite{guo2017countering}, and Liang et al.~\cite{liang2018detecting}. We also discuss the use of these operations to detect adversarial images on the basis of statistics (section~\ref{subsec:detect}) and to correct their labels as classified by the targeted object recognition DNNs (section~\ref{subsec:correct}).

\begin{figure*}[th!]
\centering
\includegraphics[width=165mm]{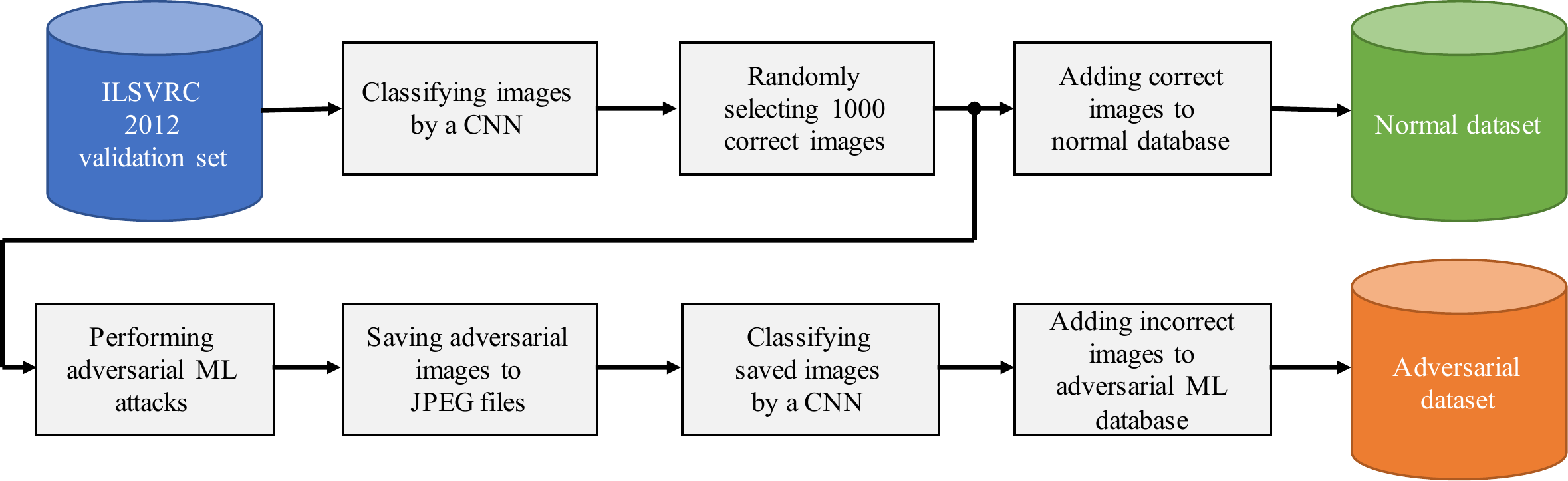}
\caption{Overview of data generation procedure using a CNN. The same procedure was used with VGG-16, VGG-19, ResNet-18, and ResNet-50 networks in order to create a complete normal dataset (for normal images) and an adversarial dataset (for adversarial images).}
\label{fig:procedure}
\end{figure*}

\section{Creating Adversarial Database}
\label{sec:create}
\subsection{Overview}
We used images from the ILSVRC 2012 validation set~\cite{ILSVRC15} (which has labeled ground truths) to generate the adversarial images used in our experiments. For the object recognition CNNs, we used pre-trained VGG-16 and -19 networks (as proposed by the Visual Geometry Group)~\cite{simonyan2014very} and pre-trained ResNet-18 and -50 networks (as proposed by Microsoft Research)~\cite{he2016deep}, implemented using the PyTorch framework~\cite{paszke2017automatic}. We used the Pillow library\footnote{https://pillow.readthedocs.io/en/stable} for image processing and the FoolBox library (version 1.8.0)~\cite{rauber2017foolbox} for adversarial image generation. Twelve commonly used methods (Table~\ref{tab:advml_atk}) were used to perform targeted and non-targeted adversarial attacks.

An overview of the procedure we used is shown in Fig.~\ref{fig:procedure}. We used each CNN to classify 5 million images from the ILSVRC 2012 validation set. We then randomly selected 1000 images per CNN that were correctly classified (the ground-truth labels were among the predicted top-5 results). Only 1000 image per CNN were selected due to the time required to create adversarial images from them. The selected images were added to the normal dataset. We then performed the 12 adversarial attacks listed in Table~\ref{tab:advml_atk} on the selected images and selected misclassified ones (those for which the predicted top-5 results did not contain the previously predicted labels). The Pillow library was used to save them as JPEG files with a quality factor of 100 to ensure that the attack was close to a real-world scenario since adversarial noise can be lost when saving adversarial images to a file. We then loaded the saved images and used the same CNN to classify them again to ensure that they were still misclassified. The misclassified adversarial images were then added to the adversarial dataset. This procedure was also used with VGG-16, VGG-19, ResNet-18, and ResNet-50 networks.

\subsection{Creating Adversarial Images and Database}
For targeted attacks, we used the limited-memory Broyden-Fletcher-Goldfarb-Shanno (L-BFGS) method proposed by Szegedy et al.~\cite{szegedy2013intriguing}, the basic iterative method (BIM), which uses L-infinity, the projected gradient descent (PGD) method described by Kurakin et al.~\cite{kurakin2016adversarial}, and the L1- and L2- versions of BIM (L1-iter and L2-iter) implemented in FoolBox~\cite{rauber2017foolbox}. Since the ILSVRC 2012 validation set has 1000 labels, the target label index for each adversarial image was its 100-step-right-shifted top-1 predicted label index (calculated using the target CNN). Modular operation was used to ensure the shifted label index was in the range $[0, 1000)$. The attack criteria for the target class probability was 99\%.

For non-targeted attacks, we used the basic gradient attack method implemented in FoolBox~\cite{rauber2017foolbox}, the fast gradient signed method (FGSM) proposed by Goodfellow et al.~\cite{goodfellow2014explaining}, the Deep Fool method proposed by Moosavi-Dezfooli et al.~\cite{moosavi2016deepfool}, the Newton method proposed by Jang et al.~\cite{jang2017objective}, the ADef method proposed by Alaifari et al.~\cite{alaifari2018adef}, the Saliency Map method proposed by Papernot et al.~\cite{papernot2016limitations}, and the attack method proposed by Carlini and Wagner~\cite{carlini2017towards}. Since these methods are non-targeted attacks, the attack criterion was to change the predicted top-5 labels so that they were different from the original predicted top-1 labels.

\begin{table}[th!]
\caption{Number of successful adversarial images created using FoolBox library~\cite{rauber2017foolbox} on VGG-16, VGG-19, ResNet-18, and ResNet-50 networks.}
\label{tab:advml_atk}
\adjustbox{max width=\columnwidth}{%
\centering
\begin{tabular}{l|c|c|c|c}
\multicolumn{1}{c|}{\textbf{Method}} & \textbf{VGG-16} & \textbf{VGG-19} & \textbf{ResNet-18} & \textbf{ResNet-50} \\ \hline
\textbf{\underline{Targeted attacks:}} & & & & \\
L-BFGS~\cite{szegedy2013intriguing} & 618 & 623 & 348 & 229 \\
BIM~\cite{kurakin2016adversarial} & 693 & 711 & 531 & 431 \\
PGD~\cite{kurakin2016adversarial} & 651 & 678 & 484 & 348 \\
L1-iter~\cite{rauber2017foolbox} & 661 & 591 & 680 & 513 \\
L2-iter~\cite{rauber2017foolbox} & 820 & 746 & 722 & 583 \\ \hline
\textbf{\underline{Non-targeted attacks:}} & & & & \\
Gradient~\cite{rauber2017foolbox} & 654 & 548 & 590 & 517 \\
FGSM~\cite{goodfellow2014explaining} & 509 & 450 & 451 & 379 \\
Deep Fool~\cite{moosavi2016deepfool} & 707 & 671 & 658 & 604 \\
Newton~\cite{jang2017objective} & 571 & 505 & 558 & 425 \\
ADef~\cite{alaifari2018adef} & 1 & 4 & 1 & 1 \\
Saliency Map~\cite{papernot2016limitations} & 102 & 86 & 114 & 71 \\
Carlini-Wagner~\cite{carlini2017towards} & 427 & 361 & 401 & 299
\end{tabular}}
\end{table}

As shown by the results in Table~\ref{tab:advml_atk}, the VGG networks were generally more vulnerable to targeted attacks than the ResNet networks, resulting in a larger numbers of misclassified images. The modified BIM attack using the L2 distance (L2-iter) was the most effective attack overall. The non-targeted attacks were more difficult to carry out since they needed to change the top-5 labels so that they did not include the current top-1 labels. Among them, Deep Fool was the most successful while the ADef attack was the least successful overall, producing only one or four adversarial images for each network. The Saliency Map attack also had limited success, with around 100 adversarial images for each network.

We mixed together the 1000 normal images from each classifier and the created adversarial images and divided them into training (train), development (dev), and evaluation (eval) sets at an approximate ratio of 7.0:1.5:1.5, as detailed in Table~\ref{tab:database}. The train set was used for training, the dev set was used to select the model, and the eval set was used to test the classifiers. We ensured that the normal images and their adversarial versions in the three sets did not overlap so that the classifier would not remember the training images. We did not focus on detecting adversarial images created by unseen adversarial attacks, so the images from each attack had equal probabilities of appearing in the three sets. The number of adversarial images was more than five times that of normal images.

\begin{table}[th!]
\caption{Details of normal and adversarial image datasets, which were divided into training, development, and evaluation sets.}
\label{tab:database}
\centering
\begin{tabular}{l|cccc}
\multicolumn{1}{c|}{\textbf{Dataset}} & \textbf{Normal} & \textbf{Adversarial} & \textbf{Total} & \textbf{Ratio} \\ \hline
Train & 2,800 & 16,110 & 18,910 & 1:5.75 \\
Dev & 600 & 3,092 & 3,692 & 1:5.15 \\
Eval & 600 & 3,124 & 3,724 & 1:5.21
\end{tabular}
\end{table}

\section{Effects of Image Processing Operations on Normal and Adversarial Images and Their Applications}
\label{sec:effects}
In section~\ref{subsec:effects}, we describe the effects of applying four image processing operations to images (normal and adversarial) on the top-5 classification results for VGG-16, VGG-19, ResNet-18, and ResNet-50 networks. Differences in behavior between normal and adversarial images were identified, and these differences could be useful for detecting (section~\ref{subsec:detect}) and correcting adversarial images (section~\ref{subsec:correct}).

\subsection{Effects of Image Processing Operations on Normal and Adversarial Images}
\label{subsec:effects}
We extended the ideas of Xu et al.~\cite{xu2017feature}, Guo et al.~\cite{guo2017countering}, and Liang et al.~\cite{liang2018detecting} by measuring the change in the top-5 labels after applying four image processing operations with various parameters (listed below) to normal and adversarial images.
\begin{itemize}
\item JPEG compression with quality $\in$ \{100, 95, 90, 85, 80, 75, 70, 65, 60, 55, 50, 45, 40, 35, 30, 25\}.
\item Gaussian blur with kernel size $\in$ \{2, 3, 4, 5\}.
\item Clockwise image rotation with angle $\in$ \{1\textdegree, 2\textdegree, 3\textdegree, 4\textdegree, 5\textdegree, 6\textdegree, 7\textdegree, 8\textdegree\} and without reversing back.
\item Image scaling with scale $\in$ \{0.75, 0.8, 0.85, 0.9, 0.95, 1.05, 1.1, 1.15, 1.2, 1.25\} and without reversing back.
\end{itemize}

The image operations were done using Pillow version 6.1.0. Some of the results for ResNet-18 are shown in Table~\ref{tab:img_proc}. Similar behaviors were also observed for VGG-16, VGG-19, and ResNet-50. The JPEG compression with quality 100 changed the top-5 labels for both the normal and adversarial images, however its effect on adversarial images was clearer. Reducing image quality by increasing the compression ratio greatly reduced the number of misclassified adversarial images and slightly increased that of misclassified normal images. A large increase in the compression ratio increased the misclassification rate for normal images. The results for scaling, Gaussian blur, and rotation were similar to those for compression. One result in particular should be noted: the number of misclassified normal images after applying $3 \times 3$ Gaussian blur kernel was higher than after applying the other operations, which is not good for our purposes.

\begin{table*}[th!]
\caption{Number of top-5 misclassified images for ResNet-18 network from both normal and adversarial datasets before and after applying image processing operations.}
\label{tab:img_proc}
\adjustbox{max width=\textwidth}{%
\centering
\begin{tabular}{l|c|ccccc|cccccc|c|cc}
\multicolumn{1}{c|}{\multirow{2}{*}{\textbf{Attack}}} & \multicolumn{1}{c|}{\multirow{2}{*}{\textbf{Original}}} & \multicolumn{5}{c|}{\textbf{JPEG Compression}} & \multicolumn{6}{c|}{\textbf{Scaling}} & \textbf{Gaussian} & \multicolumn{2}{c}{\textbf{Rotation}} \\ \cline{3-16} 
\multicolumn{1}{c|}{} & \multicolumn{1}{c|}{} & \textbf{100} & \textbf{80} & \textbf{60} & \textbf{40} & \textbf{20} & \textbf{0.75} & \textbf{0.85} & \textbf{0.95} & \textbf{1.05} & \textbf{1.15} & \textbf{1.25} & \textbf{3 $\times$ 3} & \textbf{\begin{tabular}[c]{@{}c@{}}2\textdegree\end{tabular}} & \textbf{\begin{tabular}[c]{@{}c@{}}5\textdegree\end{tabular}} \\ \hline
Normal images & 0 & 6 & 29 & 50 & 59 & 108 & 93 & 44 & 34 & 26 & 52 & 35 & 329 & 54 & 75 \\ \hline
{\ul \textbf{Targeted attack:}} & & & & & & & & & & & & & & & \\
L-BFGS~\cite{szegedy2013intriguing} & 348 & 341 & 41 & 50 & 58 & 67 & 88 & 58 & 44 & 59 & 68 & 42 & 63 & 61 & 92 \\
BIM~\cite{kurakin2016adversarial} & 531 & 529 & 52 & 54 & 69 & 88 & 103 & 66 & 55 & 65 & 84 & 47 & 87 & 72 & 101 \\
PGD~\cite{kurakin2016adversarial} & 484 & 479 & 45 & 50 & 61 & 82 & 102 & 60 & 53 & 61 & 77 & 42 & 77 & 65 & 95 \\
L1-iter~\cite{rauber2017foolbox} & 680 & 676 & 60 & 62 & 61 & 81 & 100 & 71 & 65 & 80 & 92 & 60 & 93 & 91 & 115 \\
L2-iter~\cite{rauber2017foolbox} & 722 & 720 & 66 & 59 & 72 & 88 & 107 & 72 & 69 & 88 & 99 & 60 & 100 & 99 & 123 \\ \hline
{\ul \textbf{Non-targeted attack:}} & & & & & & & & & & & & & & & \\
Gradient~\cite{rauber2017foolbox} & 590 & 584 & 107 & 76 & 66 & 95 & 93 & 70 & 75 & 115 & 122 & 86 & 108 & 105 & 155 \\
FGSM~\cite{goodfellow2014explaining} & 451 & 449 & 109 & 87 & 76 & 101 & 99 & 67 & 79 & 125 & 124 & 93 & 95 & 110 & 158 \\
Deep Fool~\cite{moosavi2016deepfool} & 658 & 645 & 83 & 55 & 51 & 72 & 60 & 51 & 59 & 74 & 72 & 46 & 59 & 80 & 50 \\
Newton~\cite{jang2017objective} & 558 & 552 & 118 & 61 & 60 & 74 & 62 & 61 & 64 & 90 & 82 & 48 & 82 & 105 & 68 \\
ADef~\cite{alaifari2018adef} & 1 & 1 & 1 & 1 & 1 & 0 & 0 & 0 & 0 & 0 & 0 & 0 & 0 & 0 & 0 \\
Saliency Map~\cite{papernot2016limitations} & 114 & 109 & 42 & 41 & 40 & 54 & 59 & 40 & 28 & 36 & 38 & 27 & 33 & 43 & 44 \\
Carlini-Wagner~\cite{carlini2017towards} & 401 & 391 & 59 & 50 & 48 & 75 & 68 & 43 & 42 & 59 & 59 & 45 & 38 & 67 & 49
\end{tabular}}
\end{table*}

\begin{table*}[th!]
\caption{Accuracy (in \%) of each classifier using counting feature (count) and difference feature (diff.) on eval set.}
\label{tab:dec_img}
\adjustbox{max width=\textwidth}{
\centering
\begin{tabular}{l|cc|cc|cc|cc|cc|cc}
\multicolumn{1}{c|}{\multirow{2}{*}{\textbf{Classifier}}} & \multicolumn{2}{c|}{\textbf{JPEG}} & \multicolumn{2}{c|}{\textbf{Scaling}} & \multicolumn{2}{c|}{\textbf{Gaussian Blur}} & \multicolumn{2}{c|}{\textbf{Rotation}} & \multicolumn{2}{c|}{\textbf{JPEG + Scaling}} & \multicolumn{2}{c}{\textbf{All}} \\ \cline{2-13} 
 \multicolumn{1}{c|}{} & \textbf{Count} & \textbf{Diff.} & \textbf{Count} & \textbf{Diff.} & \textbf{Count} & \textbf{Diff.} & \textbf{Count} & \textbf{Diff.} & \textbf{Count} & \textbf{Diff.} & \textbf{Count} & \textbf{Diff.} \\ \hline
SVM (SVC)~\cite{cortes1995support} & 90.98 & 91.92 & 92.56 & 92.32 & 87.35 & 87.35 & 89.90 & 89.39 & 92.86 & \textbf{94.04} & 93.39 & \textbf{94.20} \\
Random forest~\cite{ho1995random} & 90.57 & 89.82 & 91.08 & 91.27 & 86.87 & 86.04 & 88.24 & 88.59 & 92.16 & 91.94 & 92.35 & 92.35 \\
LDA~\cite{duda1973pattern} & 89.98 & 90.36 & 91.97 & 91.62 & 85.77 & 87.35 & 89.15 & 89.29 & 92.86 & 92.67 & 92.21 & 92.86 \\
MLP~\cite{ruck1990multilayer} & 91.25 & 90.41 & 92.32 & 92.35 & 86.95 & 87.38 & 89.66 & 89.02 & \textbf{93.56} & 92.19 & \textbf{93.37} & 92.29
\end{tabular}}
\end{table*}

\subsection{Detecting Adversarial Images using Statistical Features}
\label{subsec:detect}
As mentioned above, the differences in behavior between normal and adversarial images for the top-5 labels when the four image processing operations are applied may be useful for detecting adversarial images. We thus propose using two features for detecting adversarial images: a counting feature and a differences feature. First, we define three variables.

\begin{itemize}
\item $L = (a, b, c, d, e)$: top-5 label for image $I$ (normal or adversarial) predicted using a CNN before applying image processing operation $i$ (e.g., JPEG compression with a quality of 80 or 5\textdegree clockwise rotation).
\item $L_i = (a_i, b_i, c_i, d_i, e_i)$: top-5 label for an image after applying image processing operation $i$.
\item $n$: total number of image processing operations (38 in our experiments).
\end{itemize}

\subsubsection{Counting feature}
Let us call $C(a)$ the number of occurrences of label $a \in L = (a, b, c, d, e)$ at the first position in an ordered top-5 label set $\{L_i | i = 1..n\}$ and $ \mathbbm{1}(.)$ the counting function. $C(a)$ is defined as
\begin{equation}
    C(a) = \sum_{i=1}^{n} \mathbbm{1}(a_i = a).
\end{equation}

The same equation is used for $b$, $c$, $d$, and $e$ at the second, third, four, and fifth positions, respectively. Therefore, the features of each image are $\{C(a), C(b), C(c), C(d), C(e)\}$.

\subsubsection{Differences feature}
Let us call $\Delta(a, a_i)$ the binary differential function used to measure the difference between two labels, $a$ and $a_i$:
\begin{equation}
    \Delta(a, a_i) = 
    \begin{cases}
        1,& \text{if } a_i\neq a\\
        0,& \text{if } a_i = a.\\
    \end{cases}
\end{equation}

The differences feature derived from input image $I$ can be expressed as the set 
$$\{(\Delta(a, a_i), \Delta(b, b_i), \Delta(c, c_i), \Delta(d, d_i), \Delta(e, e_i)) | i = 1..n\}.$$

\subsubsection{Detecting adversarial images using image-processing-based features}
Using the two features introduced above, we evaluated the performance of detectors using one of the four classifiers: the C-support vector classification (SVC) version of the support vector machine (SVM) classifier~\cite{cortes1995support}, the random forest classifier~\cite{ho1995random} with 100 estimators and a maximum depth of 2, the linear discriminant analysis (LDA) classifier~\cite{duda1973pattern}, and the multiple layer perception (MLP) classifier ~\cite{ruck1990multilayer}. All of them were implemented in the scikit-learn library version 0.21.3\footnote{https://scikit-learn.org/stable}.

As shown in Table~\ref{tab:dec_img}, the differences feature produced slightly higher accuracies than the counting feature (which was smaller than the differences feature). Among the individual image processing operations, the scaling one achieved the highest accuracy for all detectors. Also as shown in the table, the combination of JPEG compression and scaling and the combination of all operations resulted in higher accuracy than using any of them separately. However, these combinations also increased the feature size and more classification operations were required of the CNNs (VGG-Nets or ResNets) to produce those features. For the counting feature, using the MLP-based detector on the features from JPEG compression and the scaling operation resulted in the highest accuracy (93.56\%) while for the differences feature, using the SVM-based detector on all features from all image processing operations resulted in the highest accuracy (94.20\%).

\subsection{Correcting Adversarial Images}
\label{subsec:correct}
Given that the image processing operations substantially reduced the number of misclassified adversarial images, they could also be useful in restoring the original labels of the images in addition to classifying them. Those operations helped mitigate the adversarial noise while only slightly affecting the normal images. Since there are several kinds of adversarial attacks, we focused on correcting noise-based adversarial images. The correction task is usually performed after adversarial image detection. In this section, we introduce our proposed correction method and the experiment results.

\subsubsection{Proposed correction method}
Let us call $\mathbf{S} = \{L_i | i = 1..n\}$ the set of top-5 labels acquired by applying $n$ image processing operations to image $I$. We calculated the frequencies of every label in $\mathbf{S}$ and identified the five labels with the highest frequencies. These labels were the corrected top-5 ones.

\subsubsection{Evaluation}
Since there were false positive inputs, i.e., normal images misclassified as adversarial images, we tested our correction method on both normal and adversarial images. We used our entire database for testing. For the image processing operations, we used JPEG compression, scaling, and their combination. As shown in Table~\ref{tab:correct}, JPEG compression had better performance than scaling operation for the adversarial images, and their combination produced the best performance. Only 1.88\% of the normal images was misclassified after ``correction'' while 89.91\% of the adversarial images were corrected. If we performed this correction after detection of adversarial images, 98.12\% of the false positive inputs (normal images misclassified as adversarial ones) would also be corrected.

\begin{table}[th!]
\caption{Percentage of corrected top-5 classifications after applying proposed correction method to both normal and adversarial images.}
\label{tab:correct}
\centering
\begin{tabular}{l|c|c}
\multicolumn{1}{c|}{\textbf{Operation(s)}} & \textbf{Normal Images} & \textbf{Adversarial Images} \\ \hline
JPEG compression & 96.30 & 88.63 \\
Scaling & 97.82 & 85.69 \\
\textbf{Both} & \textbf{98.12} & \textbf{89.91}
\end{tabular}
\end{table}

\section{Summary and Future Work}
The statistical-based detector using image processing operations demonstrated its ability to detect adversarial images with high accuracy. The corresponding correction method to restore the original labels of adversarial images can be used together with the proposed detectors to improve the performance of DNNs under adversarial attack. Future work will address more adversarial attacks with multiple noise strengths on larger and more diverse databases along with reducing the number of image processing operations in order to reduce computational expense.

\section*{Acknowledgements}

This research was supported by JSPS KAKENHI Grants JP16H06302, JP17H04687, JP18H04120, JP18H04112, JP18KT0051, JP19K22846, and by JST CREST Grant JPMJCR18A6, Japan.

\bibliographystyle{IEEEbib}
\bibliography{refs}

\end{document}